\documentclass[conference]{IEEEtran}
\IEEEoverridecommandlockouts
\usepackage{times}
\usepackage{latexsym}
\usepackage{xcolor}
\usepackage{caption}
\usepackage{subcaption}
\usepackage{bm}
\usepackage{multirow}
\usepackage{microtype}
\usepackage{hyperref}
\usepackage{cleveref}
\usepackage{eso-pic}

\usepackage{tcolorbox}
\tcbuselibrary{skins}
\makeatletter
\def\blfootnote{\gdef\@thefnmark{}\@footnotetext}
\makeatother
\newcommand{\hand}[1] {
    \begin{tcolorbox}[skin=enhanced,
                    width=3in,
                    colback=white,
                    fontlower=\sffamily,
                    fontupper=~\rmfamily,
                    middle=0mm,
                    center
                    ]%%
    \includegraphics[width=0.46cm]{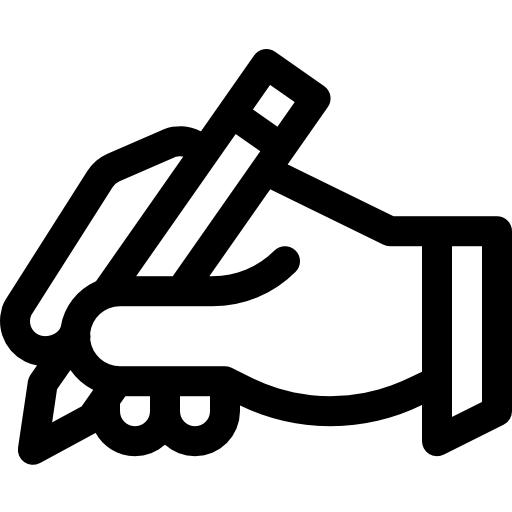}
    #1
    \end{tcolorbox}
}

\newcommand{\aws}[1] {
    \begin{tcolorbox}[skin=enhanced,
                    width=2.65in,
                    colback=white,
                    fontlower=\sffamily,
                    fontupper=~\rmfamily,
                    middle=0mm,
                    center
                    ]%%
    \includegraphics[width=0.46cm]{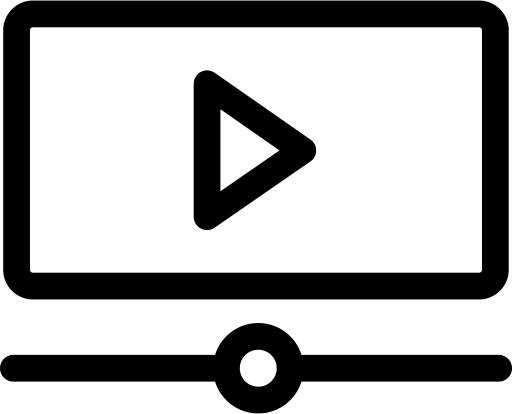}
    #1
    \end{tcolorbox}
}

\usepackage{graphicx}
\newcommand{\eg}{e.\,g.\,, }
\newcommand{\ie}{i.\,e.\,, }

\newcommand{\et}{{et al.\,}}
\newcommand{\cf}{{cf.\,}}
\newcommand{\musec}{MuSe-CaR}
\newcommand{\city}{Citysearch}

\newcommand{\pthresh}{$p_{t}$}

\newcommand{\lda}{LDA}

\newcommand{\hdbscan}{HDBSCAN}

\newcommand{\graph}{GraphTMT}
\newcommand{\wip}{WIP}
\newcommand{\nsf}{NSF}

\usepackage{dsfont}

\usepackage[absolute,showboxes]{textpos}

\TPMargin{5pt}

\newcommand{\copyrightstatement}{
    \begin{textblock}{0.84}(0.08,0.017)    % tweak here: {box width}(leftposition, rightposition)
         \noindent
         \footnotesize
         \copyright 2021 IEEE.  Personal use of this material is permitted.  Permission from IEEE must be obtained for all other uses, in any current or future media, including reprinting/republishing this material for advertising or promotional purposes, creating new collective works, for resale or redistribution to servers or lists, or reuse of any copyrighted component of this work in other works.
    \end{textblock}
}

\begin{document}

\title{GraphTMT: Unsupervised Graph-based Topic Modeling from Video Transcripts}

%\AddToShipoutPicture*{\small \sffamily\raisebox{1.2cm}{\hspace{1.8cm}978-1-4799-8641-5/15/\$31.00@2015 IEEE}}
 
\author{\IEEEauthorblockN{1\textsuperscript{st} Jason Thies}
\IEEEauthorblockA{\textit{Social Computing} \\
\textit{Technical University of Munich}\\
Munich, Germany \\
jason.thies@tum.de}
\\
\IEEEauthorblockN{3\textsuperscript{rd} Bj{\"o}rn W.\ Schuller}
\IEEEauthorblockA{\textit{GLAM} \\
\textit{Imperial College London}\\
London, United Kingdom \\
bjoern.schuller@imperial.ac.uk}
\and
\IEEEauthorblockN{1\textsuperscript{st} Lukas Stappen\thanks{$^{1}$JT and LS contributed equally to this work.}}
\IEEEauthorblockA{\textit{EIHW} \\
\textit{University of Augsburg}\\
Augsburg, Germany \\
stappen@ieee.org}
\\
\IEEEauthorblockN{4\textsuperscript{th} Georg Groh}
\IEEEauthorblockA{\textit{Social Computing} \\
\textit{Technical University of Munich}\\
Munich, Germany \\
grohg@mytum.de}
\and
\IEEEauthorblockN{2\textsuperscript{nd} Gerhard Hagerer}
\IEEEauthorblockA{\textit{Social Computing} \\
\textit{Technical University of Munich}\\
Munich, Germany \\
ghagerer@mytum.de}

}

\copyrightstatement
\maketitle

\begin{abstract}
To unfold the tremendous amount of multimedia data uploaded daily to social media platforms, effective topic modeling techniques are needed. Existing work tends to apply topic models on written text datasets. In this paper, we propose a topic extractor on video transcripts. Exploiting neural word embeddings through graph-based clustering, we aim to improve usability and semantic coherence. Unlike most topic models, this approach works without knowing the true number of topics, which is important when no such assumption can or should be made. Experimental results on the real-life multimodal dataset MuSe-CaR demonstrates that our approach GraphTMT extracts coherent and meaningful topics and outperforms baseline methods. Furthermore, we successfully demonstrate the applicability of our approach on the popular Citysearch corpus.

%%\vspace{-.8em}
%\keywords{topic modeling  \and graph connectivity  \and transcripts \and k-components.}
\end{abstract}


\section{Introduction}

%%
% Paragraph Content: general introductory (motivational) information   
%%
Hundreds of hours of videos are uploaded to YouTube every minute, enabling studies in various fields of research. For example, educational information on cancer treatment~\cite{basch2017content} and hearing aids~\cite{manchaiah2020content} are studied in health-care, the influence on election campaigns in social sciences~\cite{gueorguieva2008voters}, and large-scale multimodal sentiment in multimodal machine learning~\cite{wollmer2013youtube, morency2011towards, stappen2021multimodal, stappen2021challenge, stappen2021estimation}. 
% example in what way they are analysed e.g. visual?
For these approaches, researchers closely examine the videos for collection, labelling, and analysis, whereby visual patterns and metadata, \eg authorship, can be exploited. Nowadays, also transcripts -- automatically created by YouTube -- are available \cite{youtubeCaptions}. Since text is the most meaningful modality to understand contextual information, effective computer-assisted text analysis methods are needed.
%This raises the question of how this new source can be harnessed to gain a richer contextual understanding of online videos. 

%%
% topic model intro + applications + example
%%
Topic models that structure information into theme distributions have existed for many years. It has been performed on a range of different texts, including online social network data~\cite{curiskis2020evaluation, Missier-2016-Tracking,prabhakar2020informational}, journals~\cite{jacobi2015Quantitative}, and transcripts~\cite{basu2016fuzzy, morchid2013a}. %However, the natural language of transcripts come with its own challenges, being inherently colloquial,

Given a transcript snippet: ``\textit{It comes with four turbochargers on [and] has an aught [$\Rightarrow$ naught] to 62 [$\Rightarrow$ 60] time of just 5.2 seconds and [...]}'', a typical two-way topic modeling procedure \textit{first}, extracts the aspect terms \eg ``turbochargers'', \textit{second}, clusters the aspects into coherent topic clusters \eg ``motorisation'' = \{``turbochargers'', ``engine'', ...\}.  
%%
% transcript challenges
%%
Automatic transcripts, however, bring unique challenges. Transcripts often have errors like missing words (``\textit{and}''), incorrect (``\textit{62}'' $\Rightarrow$ ``\textit{60}''), and similar sounding words (``\textit{aught}'' $\Rightarrow$ ``\textit{naught}'') due to erroneous speech-2-text processing. 
%suitable models for this source are barely discussed in the literature. 

% eed to be effective on such colloquial data. 
% Natural language is inherently unstructured. The different forms of natural language: formal, colloquial, and verbal encompass individual challenges that need to be addressed~\cite{lo2017Multilingual}.

%%
% Paragraph Content: what is a topic modeling + video transcripts 
%%
% Topic modelling is a well-known challenge in the information retrieval community.
% The 
%In this case, given a set of utterances, ,....
%transcripts, a topic model can discover the latent semantic structure (the topics) discussed in a video, allowing the user to get a high-level understanding of the videos without the need to watch them. 
%An effective topic model approach to transcripts must respect the word error rate of the speech recognition system \cite{basu2016fuzzy, morchid2013a} used to transcribe the videos. Word error rate is a standard metric in speed recognition used to evaluate the accuracy of transcripts.
%In that regard, there is a recent research trend showing promising applications of video transcripts to provide meaningful semantic features for multimodal video indexing \cite{8387512} and summarization \cite{8732253} or to detect which video product reviews are actually perceived as being helpful \cite{park2020makes}. The question of how to optimize topic coherence on this type of data is not fully answered yet.

Video transcripts are an emerging data domain, however, the explicit use for topic modeling is understudied~\cite{morchid2013a,basu2016fuzzy,das2019a}. To broaden the perspective on this medium more evaluation and new approaches are needed. Recently, graph connectivity showed promising results on extracting topic from news articles~\cite{altuncu-2021-graph}.
Compared to other methods~\cite{sia-etal-2020-tired,curiskis2020evaluation,radu-2020-clustering}, the number of expected topics does not have to be explicitly determined a priori. In addition, graph modelling research has gained momentum in several areas, such as text classification~\cite{yao2019graph} and video retrieval~\cite{chen2020fine}.

In this work, we propose a \textit{Graph-based Topic Modeling approach for Transcripts} (GraphTMT). For benchmarking, we base our evaluation on \textit{a)} a problem-specific multimedia dataset of car reviews, \musec~\cite{stappen2021multimodal}, and \textit{b)} the popular written-text dataset \city~\cite{mcAuley2013hidden}. \musec~is one of the largest state-of-the-art video datasets for multimodal sentiment analysis research, containing almost 40 hours of video footage and transcripts of car reviews. The reported word error rate of the automatic transcript is estimated around 28\,\%~\cite{stappen2021multimodal}. To the best of our knowledge, studies on topic extraction have only been conducted in a supervised fashion ~\cite{stappen2020muse, stappen2021sentiment, stappen2021musebox} on this corpus. Furthermore, \city{} is utilised to evaluate the applicability of our approach to other datasets. It covers written reviews from restaurant visits and is often featured for the task of aspect and topic modeling in previous works~\cite{mcAuley2013hidden, brody2010an, he2017unsupervised}.

%%
% research contributions
%%

%This study evaluates the general performance of graph-based topic modeling against common topic modeling approaches.  , and it is the first time this is done in an unsupervised graph-based clustering procedure. 
%%
% Research Questions
%%
% In this paper, we aim to answer the following research questions with our proposed graph connectivity approach: 
% \begin{enumerate}
% \item Can meaningful and semantically coherent topics be extracted from transcribed video reviews?
% %\item Does it outperform baseline approaches while maintaining comparable execution times?
% \item Is the approach generalizable to written reviews?
% \end{enumerate}
Our contributions are as follows: We propose a novel graph-based approach for topic modeling for the emerging use case of video transcripts.
It is the first time, an unsupervised extraction model is applied to a large-scale, noisy MuSe-CaR dataset packed with typical mistakes of automatic speech-to-text.
The performance is extensively benchmarked on this dataset against conventional methods. 
Here, the semantic consistency of the topics is evaluated by assessing a common coherence measure. 
Furthermore, for a more human-centred evaluation approach of the results and to determine the semantic validity, we conduct a structured word intrusion user study with 31 subjects.
Finally, we evaluate the coherence of our approach on a standard topic modeling dataset of product reviews to assess the potential for other use cases.
% What benefits will it have for future research?
Our results show that GraphTMT outperforms conventional methods on the MuSe-CaR datasets. %  We hope to pave the way for a more {was sind graph-ansätze?} on this understudied data type. 
For reproducibility, this paper is adjoined with a public Git repository\footnote{Our code can be found at \textit{\url{https://github.com/JaTrev/unsupervised_graph-based}}}. 
%The paper's structure is as follows: First, our approaching is explained in detail, followed by a description of the datasets. Next, we present the evaluation results and, finally, discuss the results.

%%%%%%%%%%%%%%%%%%%%%%
%%%%%%%%%%%%%%%%%%%%%%
%
% Related Work
%
%%%%%%%%%%%%%%%%%%%%%%
%%%%%%%%%%%%%%%%%%%%%%
\section{Related Work}

%%
% Word Embedding Clustering - Topic Models
% word2vec, bert, attention-based embeddings...
%%
\subsection{Word Vector Based Topic Models}
Topic modeling is often performed by clustering natural language embeddings, grouping semantically similar words together to discover the semantic structure of the underlying corpus~\cite{curiskis2019evaluation, sahlgren-2020-rethinking, wang-etal-2020-empirical}. 

Curiskis~\cite{curiskis2019evaluation} compared a traditional topic modeling based on Latent Dirichlet Allocation (LDA) with clustering embedding approaches. All models were applied to Twitter and Reddit textual data. His study indicated that weighted and unweighted embedding clustering has the potential to outperform traditional approaches when using word2vec. 
% LDA braucht no clusters : jelodar2019latent 

Recently, Sahlgren~\cite{sahlgren-2020-rethinking} compared document-based topic modeling to word-based topic modeling. The word-based topic models used utilized embeddings for each prominent word, and the document-based model used document embeddings. The study showed that word-based topic modeling resulted in less or no overlap, more unique topics, and higher average topic coherence. Furthermore, Wang \et~\cite{wang-etal-2020-empirical} recently evaluated the performance of different topic modeling approaches on Twitter data, applying embedding clustering. The study indicates that more advanced models, such as BERT, do not necessarily outperform approaches on distributed embeddings.

% journals/ijon/KimKC17
% conf/acl/HeLND17
% conf/emnlp/DingNX18
% confaacl/Sridhar15a
% conf/apsipa/XingWZL14

%%
% Graph-based cluster - Topic Models
%%
\subsection{Graph-based Topic Models}
While these studies used clustering methods to create semantically related word groups, comparatively few have worked with graphs for topic extraction. This paper aims to motivate research in using graph connectivity for topic modeling. While common clustering techniques require strict hyperparameters, \eg~K-Means requires the true number of topics, K-Components~\cite{Matula1972k-components} does not. Altuncu \et~\cite{altuncu-2021-graph} used graph connectivity and document embeddings to extract topics. The graph nodes represent  documents, and the edges are weighted by the cosine similarity of the respective document pair. The study applied minimum spanning tree and community detection to extract document groups, representing the topics of the corpus. The study concluded that graph connectivity outperforms standard clustering techniques (\eg~ K-Means). Graph-based clustering approaches have been successfully utilized in various applications, e.g. in crime pattern analysis~ \cite{das2019a} and cohesive subgraphs' discovery for social networks~\cite{li2015Influential}.

%%
% Video Transcript Topic Modeling
%%
\subsection{Topic Modeling on Video Transcripts}

There are promising applications and use cases of topic modeling related approaches on YouTube video transcripts. Morchid and Linarès~\cite{morchid2013a} used LDA-based topic modeling on self-generated YouTube video transcripts to improve automatic tagging of the uploaded videos. While the overall tagging robustness improved compared to conventional approaches, absolute performance in predicting user-provided tags remained low. The authors argued that this is due to subjectivity and high word error rate of their custom speech recognition system. More recent works are based on the video transcripts provided by YouTube itself. Basu et al.~\cite{basu2016fuzzy} apply preprocessing using automatic spell checking and irrelevant word removal. They utilize LDA for soft assignment of topics to teaching videos and texts.%, and they claim classification improvements against baseline methods. 
Furthermore, latent semantic indexing, a technique related to topic modeling, has been leveraged for search indexing on YouTube transcripts \cite{10.1007/978-3-030-62362-3_9}. Despite existing topic modeling applications, to the authors´ best knowledge, there are no coherence evaluations of topic modeling technology on YouTube transcripts. Such tool would be helpful to extract opinion targets for opinion mining purposes on video product reviews in an unsupervised manner \cite{he2017unsupervised}, widely established approach on text-based product reviews. Our goal is to foster this research on publicly available video transcripts for market research purposes.

%Thus only potential topic representatives remain in the dataset, an approach used in previous studies~\cite{martin-johnson-2015-efficient, das2019a}.
%\missing{add related work from gerry}
%\missing{check Multimodal Sentiment Analysis papers}
%% missing: summary paragraph

%%%%%%%%%%%%%%%%%%%%%%
%%%%%%%%%%%%%%%%%%%%%%
%
%
%
% Approach
%
%
%%%%%%%%%%%%%%%%%%%%%%
%%%%%%%%%%%%%%%%%%%%%%
\section{Approach: \graph{} \label{sec:approach}}

\begin{figure}[t]
\centering
\includegraphics[width=\linewidth]{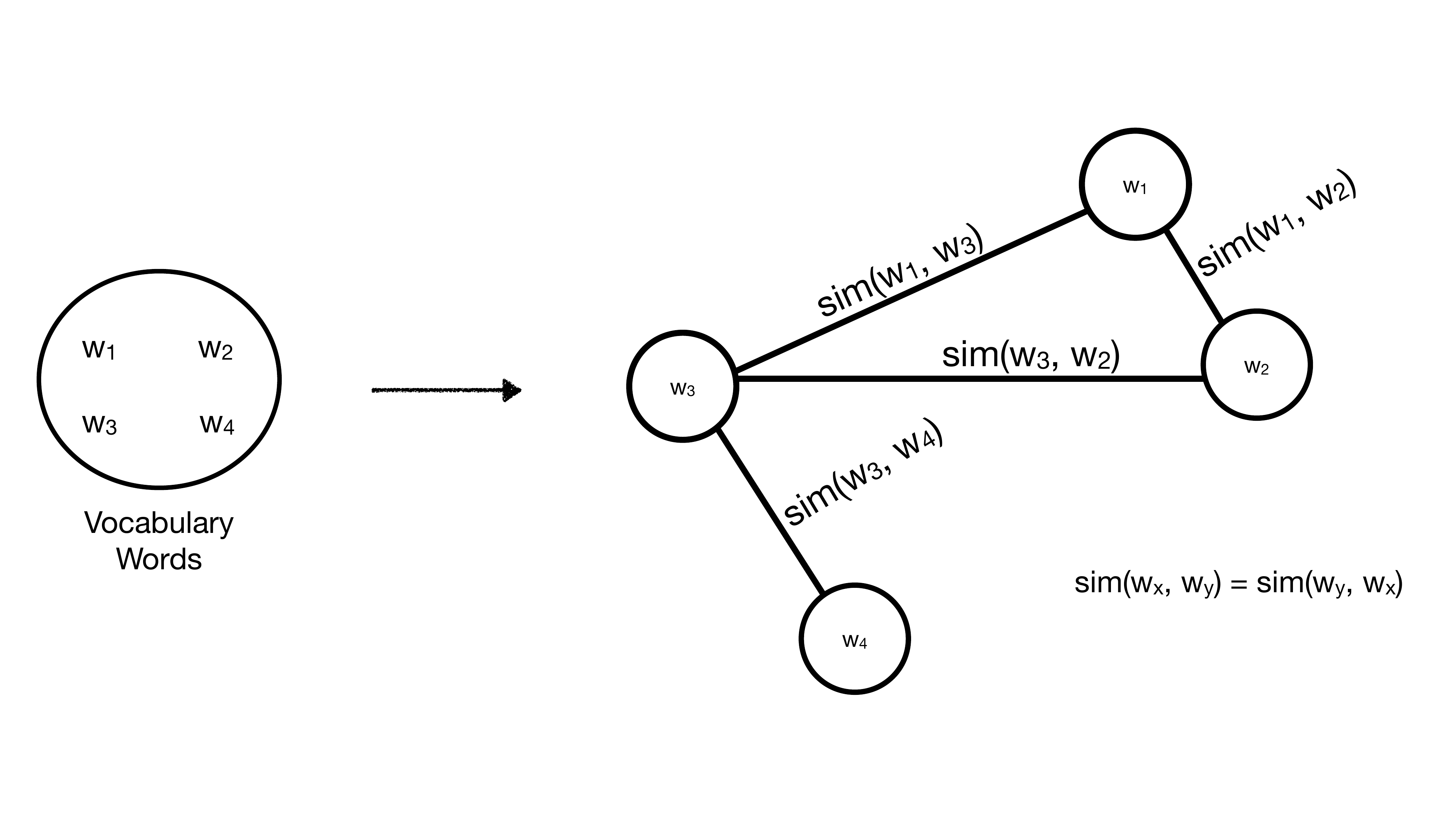}
%%\vspace{-2em}
\caption{Illustration of a word embedding graph. Each node represents a word from the vocabulary and each edge is weighted by the similarity between the adjacent nodes. The edges are undirected so that sim($w_i$, $w_j$) = sim($w_j$, $w_i$).} 
%%\vspace{-1.5em}
\label{fig:wordEmbeddingGraph}
\end{figure}

In this section, we describe our proposed graph-based topic modeling approach. The ultimate goal of \graph{} is to create and split a word embedding graph, into subgraphs based on edge connectivity. The resulting subgraphs, similar to word embedding clusters, hold semantically related words and are considered the prominent topics of the corpus.

\subsection{Word Embedding Graph \label{sec:wordEmbeddingGraph}}
%%
% Word Embedding Graph
%%
Given a set of vocabulary words $W$ ($|W|$ = $n$), a unique set of the most prominent corpus words, a word embedding graph $G$ = ($N$, $E$) is created consisting of $|N|$ $\leq$ $n$ nodes. Each node represents a vocabulary word and each undirected edge $e \in E$ is weighted by the cosine similarity score of the adjacent nodes (\cf~ \Cref{fig:wordEmbeddingGraph}). Cosine similarity is used to represent the semantic similarity embodied within the trained embeddings~\cite{levy2015Improving}. A higher cosine score indicates higher semantic similarity, while an edge weighted with a low cosine score indicates that the adjacent words are not semantically related. 

\subsection{Edge Dropping \label{sec:edge-dropping}}
By weighting the edges, low-weighted edges can be removed from the graph without disconnecting subgraphs of high semantic similarity. To extract insightful topics from the graph, \graph{} uses a percentile threshold \pthresh{} to remove low-weighted edges in $E$. 
% We also evaluated xyz, however, found this method more effective.

%The unprocessed word embedding graph $G$ is complete, every edge (vocabulary words) is adjacent to every other edge. 
%We compare two approaches to remove these, using a percentile similarity threshold (\pthresh) for the weighted edges and keeping only the top $m$ highest weighted edges (\topm). In the last step, the isolated nodes are removed from the graph. 

\subsection{Graph-based Topic Modeling \label{k-comp}}
Using the resulting (incomplete) graph, the k-component subgraphs~\cite{Matula1972k-components, Torrents2015StructuralCV, white2001fast} are calculated. A k-component is a maximal subgraph of the original graph having (at least) edge connectivity $k$, a minimum of $k$ edges must be removed from such a k-component subgraph to split it into further subgraphs. These subgraphs are inherently hierarchical; a 1-connected graph can contain several 2-component subgraphs, each of which can contain multiple 3-component subgraphs. In~\Cref{fig:wordEmbeddingGraph}, $G_{sub}$ = ($N_{sub}$, $E_{sub}$), with $N_{sub}$ = $\{w_1, w_2, w_3\}$ and $E_{sub}$ = \{sim($w_1, w_3$), sim($w_3, w_2$), sim($w_1, w_2$)\} is a 2-component subgraph of the given graph.
Each k-component subgraph represents a topic discussed in the corpus. The top $N$ representatives of each topic are selected based on node degree and node weights.

\begin{figure}[t]
  %\begin{minipage}{.47\textwidth}
    \includegraphics[width=\linewidth]{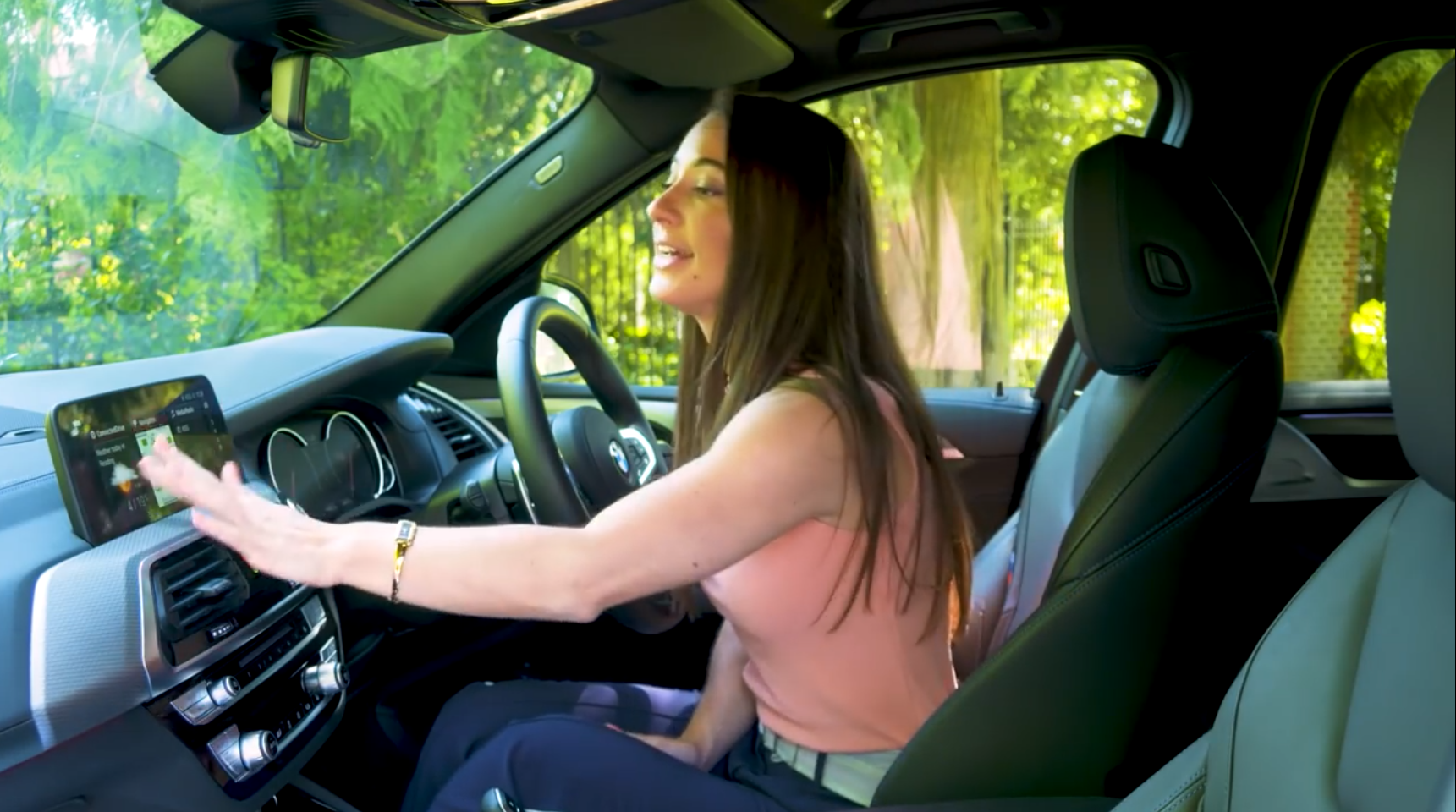}
    %\hspace{0.12cm}
  %\end{minipage} \quad
  \vspace{-.9cm}
  %\begin{minipage}{.52\textwidth}
    \aws{\scriptsize \emph{This infotainment system, though, [...] is displayed on other a 6.5 or 10.3 inch touchscreen [...] is miles ahead of anything else in its class}}
  %\end{minipage}%
  \caption{Frame from \musec{} (video id 2, 4:06) showing a \textit{User Experience} segment and corresponding transcripts.}
  %\vspace{-1em}
\end{figure}

\section{Experimental Setup}
%%\vspace{-.9cm}

%%
% datasets
%%
\subsection{Datasets}

We evaluate our method on two real-world datasets.  We focus on \musec, applying different topic modeling approaches to the unique dataset but include the \city{} corpus to demonstrate the applicability of \graph{} outside of video transcripts. 
%All clustering techniques run using implementations in scikit-learn\footnote{https://scikit-learn.org/}, we use Gensim\footnote{https://radimrehurek.com/gensim/} for implementations of \lda{} and Word2vec, and graphs are created using NetworkX\footnote{https://networkx.org}.

\paragraph{\musec:} \label{sec:musec}
The \musec{} \cite{stappen2021multimodal} is a multimodal dataset gathered in-the-wild from English YouTube videos centred around car reviews. % from several makes. 
It was created with different computational tasks in mind, allowing researchers to improve the machine's understanding of how sentiment and topics are connected. The in-the-wild aspect of \musec{} refers to the natural conditions a video is captured in. It varies in recording equipment, recording setting, and soundscapes. The audio captures ambient noises (\eg car noises), while the non-acted speech includes colloquialisms and domain-specific terms. 

For our experiments, we use a preprocessed subset of the data featuring labelled topic segments\footnote{Download MuSe-Topic: \url{https://zenodo.org/record/4134733}}, consisting of a total of 35h 39min of YouTube car review videos of approx. 90 speakers~\cite{stappen2020muse}.
Consisting of real-life opinions about different aspects of modern vehicles, the dataset allows one to apply models to a large volume of user-generated data. The corpus includes 5\,467 segments, each consisting of multiple sentences (total: $>$ 20k sentences) with an average of 54 words. Long, encapsuled utterances are typical for transcripts. 
Video segments are assigned to one of ten topics: \textit{Comfort}, \textit{Costs}, \textit{Exterior} \textit{Features}, \textit{General Information}, \textit{Handling}, \textit{Interior Features}, \textit{Performance},  \textit{Safety}, \textit{Quality} \& \textit{Aesthetic}, and \textit{User Experience}.
The transcripts are generated by the authors using automatic Amazon Transcribe speech-to-text pipelines. %These pipelines provide timestamps for each word (start and end point of an utterance) through which all words articulated within a segment can be assigned to it. 
Due to the in-the-wild factors, the error rate of the automatic transcripts is estimated to be relatively high and specified at around 28\,\% with outliers of up to 39\,\% on a subset of 10 hand-transcribed videos \cite{stappen2021multimodal}.

\paragraph{Citysearch corpus:} \label{sec:citysearchCorpus}  
Restaurant reviews from Citysearch\footnote{Download Citysearch: \url{http://www.cs.cmu.edu/~mehrbod/RR/}, accessed on 29 April 2021} have been widely used in previous works \cite{mcAuley2013hidden, brody2010an, he2017unsupervised}. Citysearch was created in 2006. The project aims to provide a better understanding of patterns in user reviews and create tools to better analyse text reviews. The corpus contains over 50\,000 restaurants reviews, written by over 30\,000 distinct users.
Ganu \et{}\cite{ganu2009beyond} manually labelled a subset of 3\,400 sentences using one of six topics: \textit{Ambience}, \textit{Anecdotes}, \textit{Food}, \textit{Miscellaneous}, \textit{Price}, and \textit{Staff}. The topic modeling approaches are evaluated based on this labeled subset.

\begin{figure}[t]
  \begin{minipage}{.5\textwidth}
    \hand{\scriptsize \emph{awful start to finish. [...] we were 1 of only 2 tables in the whole place. zero atmosphere, overpriced menu, average food [...]}} %
  \end{minipage}%
  \caption{Snippet from a review from the \city{} corpus.}
  %\vspace{-1em}
\end{figure}

%%
% Preprocessing
%%
\subsection{Preprocessing}
We begin by extracting the corpus vocabulary $W = \{w_1, w_2, ..., w_n\}$ ($|W|$ = $n$). The Natural Language Toolkit (NLTK) \cite{nltk} part-of-speech (POS) tagger is used to collect POS tags for each word. Word tags have been successfully applied in previous studies \cite{yi-2003-sentiment, das2019a}. Stop word removal is applied to the Citysearch vocabulary, due to its larger size.

% word2vec and Preprocessing
After extracting the corpus vocabulary $W$, we associate each word to a word embedding. The word2vec model \cite{Mikolov2013Efficient} is used to learn these feature vectors, using the following parameters: window size = 15, epoch = 400, hierarchical softmax, and the skip-gram word2vec model \cite{Mikolov2013Efficient}. For a fair comparison, this configuration is used in all settings.

% POS tagging
Furthermore, we run preliminary experiments on \musec{} and \city{} utilising the POS tags (\cf~\Cref{sec:approach}). The results indicated that using only nouns performs better on \musec{}, regardless of the method, while the use of all parts-of-speech tags yields slightly better results on \city{} (\cf~\Cref{sec:discuss}) which we report in the following.

%%
% Baseline approach
%%
\subsection{Baseline Approaches}
Three baseline approaches are compared with \graph: Latent Dirichlet allocation (\lda) \cite{blei_lda_2003}, K-Means~\cite{macqueen1967some}, and Hierarchical Density-Based Spatial Clustering of Applications with Noise (\hdbscan)\cite{mcinnes2017accelerated}.

\lda{} is a common topic modeling technique, using word co-occurrences to learn semantic clusters.  It uses a Dirichlet prior on the topic distribution and the topic representatives distribution.  \lda{} works with a bag-of-words (BOW) representation of the data. Each text is represented as a set of words and their cardinality, neglecting the sentence structure and context. Commonly, the BOW representation is translated into term frequency (TF) or TF-inverse document frequency (TF-IDF) matrix representation. K-Means is a common clustering technique used in topic modeling \cite{sia-etal-2020-tired, curiskis2020evaluation, radu-2020-clustering, wang-etal-2020-empirical, altuncu-2021-graph}. While \lda{} works on probability distributions of topics on the document, K-Means uses the distance between clusters. Similarly to \lda, K-Means commonly~\cite{curiskis2020evaluation} uses the TF or TF-IDF matrix representation of the data. The algorithm simultaneously divides the dataset into a number of $T_n$ clusters. The number of clusters is predefined, and the algorithm repeats two steps: an assignment and an update step. While in the assignment step, each data point is assigned to the cluster centroid based on the least squared Euclidean distance, the update step recalculates the centroids. \hdbscan{} is a hierarchical and density-based clustering technique which creates a minimum spanning tree and condenses it into smaller trees to create clusters, stopping at $C_{min}$. Unlike K-Means, \hdbscan{} allows for outliers.

%All clustering techniques run using implementations in scikit-learn\footnote{https://scikit-learn.org/}, we use Gensim\footnote{https://radimrehurek.com/gensim/} for implementations of \lda{} and Word2vec, and graphs are created using NetworkX\footnote{https://networkx.org}.

\subsection{Measures \label{sec:measures}}

The different topic modeling approaches are measured by: (1) a coherence score, (2) intra-topic assessment, and (3) a user study. 

\paragraph{(1) Coherence score} Topic coherence measures the degree of semantic similarity between topic representatives, the topic's ten most eminent words. A model's coherence score is the average of all topic scores. This study uses the $c_v$ coherence score \cite{roderCoherence15}. It is based on a sliding window approach that uses normalized pointwise mutual information (NPMI) and cosine similarity. R\"{o}der \et~\cite{roderCoherence15} studied the correlation between numerous coherence scores and human judgement and found that $c_v$ correlates best with human ratings.

\paragraph{(2) Intra-topic assessment} As coherence scores only capture the similarity between topic representatives, the intra-topic assessment compares the inferred topics with the dataset topic labels (the gold topics)~\cite{sahlgren-2020-rethinking}. It includes two measures:

\begin{itemize}

\item Topic coverage ($T_c$): how many gold topics are inferred? This is the proportion of gold topics that are included in the model's topics. A larger number indicates better gold topic representation.

\item Topic overlap ($T_o$): how much do the topics overlap? 
Each topic is given a label based on its representative, we compare these labels to find the proportion of duplicates. A small overlap indicates unique semantic structures. 

\end{itemize}

\begin{table}[t]
\centering
\resizebox{.47\textwidth}{!}{%
\begin{tabular}{lc}
\hline
\hline
\textbf{Parameter}          & \textbf{Values}                                  \\ \hline
Number of topics ($T_n$)   & {[}4; 20{]}                             \\
Document-topic density ($\alpha$)             & {[}0.1, 0.4, 0.7, 1.0, 1/$T_n${]} \\
Word-topic density ($\beta$)               & {[}0.1, 0.4, 0.7, 1.0{]}                \\
Weighting strategy & {[}TF, TF-IDF{]}                        \\
Minimum cluster size($C_{min}$) & {[}5; 30{]}  \\
Edge-connectivity ($k$) & {[}1, 2, 3{]}     \\
Edge weight threshold (\pthresh)          & {[}0.50, 0.60, 0.70, 0.80, 0.90, 0.95{]}           
\end{tabular}%
}
\caption{Parameter settings of the models \label{tab:param}}
\end{table}

\paragraph{(3) User study} Furthermore, a user study is conducted on \musec{} models to measure the human interpretability of the inferred topics. Although topic coherence is measured, the interpretability of topics does not always align with coherence scores~\cite{chang2009reading}. Our user study consists of the word intrusion task~\cite{lau2014coherence, martin-johnson-2015-efficient, chang2009reading}. Each task is composed of six words, five representatives of a single topic, and a \emph{not sure} option. The task is to find the word that represents a different topic, \ie the intruder. Given the following intrusion task: \{system, screen, diesel, menus, voice, entertainment, \emph{not sure}\}, all words besides ``diesel'' represent the same topic (infotainment). In this example, ``diesel'' is the intruder. 

A models precision defines how well the intruder detected by the participants corresponds to the true intruder. We define the Word Intrusion Precision (\wip) by the fraction of subjects that find the correct intruders,

\begin{equation}
WIP_k^m= \sum_s \mathds{1}(i^m_{k,s} = w_k^m) / S.
\end{equation}

Let $w_k^m$ be the intruder from the $k^{th}$ topic inferred by model $m$ and let $i^m_{k,s}$ be the intruder selected by participant $s$ on the $k^{th}$ topic. Let $S$ denote the number of participants in the user study. Furthermore, the fraction of subjects that chose the \emph{not sure} option (\nsf) is captured.

To reduce study complexity, each model is assessed by half of its inferred topics (chosen at random) and each topic is assessed by a single word intrusion task. Overall the study includes 31 participants, each having an upper-intermediate English level (minimum of B2 in the Common European Framework of Language Reference).

%%%%%%%%%%%%%%%%%%%%%%
%%%%%%%%%%%%%%%%%%%%%%
%
%
%
% Evaluation and Results
%
%
%%%%%%%%%%%%%%%%%%%%%%
%%%%%%%%%%%%%%%%%%%%%%

\begin{table}[t!]
\centering
\resizebox{\linewidth}{!}{%.5\textwidth
\begin{tabular}{l|c|c|cc|cccc}
\hline
\textbf{Topic Models}  & $\bm{T_n}$   & $\bm{c_v}$   &  $\bm{T_{C}}$   &   $\bm{T_{O}}$  & \textbf{\wip}    & \textbf{\nsf} \\ \hline

\lda{} ($\alpha$ = 0.10, $\beta$ = 0.70)  &  8 & .51   & .60 & .25 & .43    & .13            \\ 

K-Means{} (TF-weighted)  &  8 & .73  & \textbf{.60} & .25 & .61     & .15  \\

\hdbscan{} ($C_{min}$=6) & 11 & .63 & .60 & .4 & -  & - \\ 

\graph{} ($k$ = 1, \pthresh = 0.80) &  6  & .76   & .50 &  \textbf{.17}  & \textbf{.63}    & \textbf{.08} \\

\graph{} ($k$ = 2, \pthresh = 0.80) &  5  & \textbf{.85}   & .40 &  .20  & -    & - \\ 

\graph{} ($k$ = 3, \pthresh = 0.80) &  2  & -   & - &  0  & - & -  \\

\end{tabular}%
}
\caption{Results on \musec{} for the different topic models and five different evaluation metrics: coherence score ($c_v$), topic coverage ($T_c$), topic overlap ($T_o$), \wip, and overall \emph{not sure} fraction. Note, \hdbscan{} was not included in the user study and only one \graph model was assessed by the participants. \label{tab:musecPerformance}}
%\vspace{-2em}
\end{table}

\section{\musec{} Evaluation}

We first present the results on \musec{} followed by the performance on \city{}.
%As in comparable clustering works~\cite{he2017unsupervised, brody2010an}, we manually map each inferred topic to one of the gold standard topics according to its representative topic words.
All model parameters are optimized to maximize the topic model coherence. During the experimental process in this paper, adjustable parameters are set uniformly as shown in \Cref{tab:param}. Any model inferring less than four topics and any topic with less than 5 representatives is not considered in our evaluation.

%Furthermore, we run preliminary experiments on \musec{} and \city{} utilising the POS tags (\cf~\Cref{sec:approach}). The results indicated that using only nouns performs better on \musec{}, regardless of the method, while the use of all parts-of-speech tags yields slightly better results on \city{} (\cf~\Cref{sec:discuss}) which we report in the following.
%All experiments in this paper are run on a machine with Intel Core i5 2.30 GHz Dual-Core CPUs and 16 GB of RAM. 

\subsection{Coherence Score Comparison}

In the first set of experiments, we compare our four models (\lda, K-Means, \hdbscan, \graph) on \musec{} based on their coherence score. \Cref{tab:musecPerformance} shows the results of the best performing hyperparameters. Although the corpus has 10 gold topics, \lda{} and K-Means perform best with eight topics. The clustering-based model gets better scores using TF instead of TF-IDF. K-Means scores better than \hdbscan{} but the hierarchical clustering techniques results in more topics. Our graph-based approach results in the highest coherence score ($c_v$ = .85), achieving significant average topic coherence without specifying the number of topics ($T_n$) or the minimum size of a topic ($C_{min}$).

Furthermore, \Cref{tab:musecPerformance} shows the impact of $k$ on \graph. Increasing the edge connectivity parameter positively impacts the coherence score but at the expense of fewer topics. By increasing $k$, lower-weighted edges are removed from the graph, splitting or removing previously existing subgraphs. The new subgraphs only include the highest-weighted edges and most semantically related words. We note that \graph{} ($k$ = 3) results in only two topics, with $\geq$ 5 representatives, so it is not assessed in our experiments.

%\begin{tabular}[c]{@{}c@{}}$\alpha$ = 0.1,\\ $\beta$ = 0.7, \\ $T_n$ = 8\end{tabular} &
%\begin{tabular}[c]{@{}c@{}}$k$ = 8,\\ TF-weighted\end{tabular}  &
%\begin{tabular}[c]{@{}c@{}}$k$ = 1,\\ \pthresh = 0.5\end{tabular}  &

From these results, we can make the following observations: (1) the best performing approaches do not include 10 topics; (2) baseline approaches can be used on \musec{} to infer coherent topics; (3) clustering-based topic modeling achieves higher scores than probability-based \lda; (4) \graph{} infers the most coherent topics without the need to specify the number of topics; and (5) by increasing $k$, the overall topic coherence of \graph{} increases but $T_n$ decreases.

\subsection{Word Intrusion}

As described in \Cref{sec:measures}, the word intrusion task measures how well the inferred topics are interpretable by humans. 
\Cref{tab:musecPerformance} lists the precision results for the three best performing models (\lda, K-Means, \graph) on \musec{}. In our case, the $c_v$ score aligns well with human judgement~\cite{roderCoherence15}. The best scoring topic model (\graph) has the highest precision and the worst scoring model (\lda) has the lowest precision. Furthermore, \graph{} has the lowest \nsf{} score. These findings suggest that \graph{} results in the most interpretable topics, underlining previous coherence results.

%A models precision defines how well the intruders detected by the participants correspond to the true intruders. We define the Word Intrusion Precision (\wip) by the fraction of subjects that find the correct intruders,
%\begin{equation}
%WIP_k^m= \sum_s \mathds{1}(i^m_{k,s} = w_k^m) / S.
%\end{equation}
%Let $w_k^m$ be the intruder from the $k^{th}$ topic inferred by model $m$ and let $i^m_{k,s}$ be the intruder selected by participant $s$ on the $k^{th}$ topic. Let $S$ denote the number of participants in the user study.

\subsection{Intra-Topic Assessment}
 
The previous two sections show K-Means and \graph{} having the best topic coherence and \wip. This section looks at these two models’ topic coverage and overlap (cf. \Cref{tab:musecPerformance}). K-Means has higher topic coverage than \graph{}, but \graph{} has a lower overlap between its topics. The overlap between topics reduces when we increase the edge connectivity constraint ($k$) but at the expense of topic coverage.

The eight topics inferred by K-Means (TF-weighted) are listed in \Cref{tab:topics-muse}. Each topic is given a label, based on its topic representatives, and assigned to a gold topic. Overall, six unique gold topics can be matched ($T_c$ = 6/10) but two topics are duplicates ($T_o$ = 2/8).

\Cref{tab:topics-muse} (middle) lists the six \graph{} ($k$=1) topics. The topics include five gold topics ($T_c$ = 5/10) and one overlap ($T_o$ = 1/6). These topics can be compared to \graph{} ($k$=2) in \Cref{tab:topics-muse}. By increasing $k$, one of the two inferred \textit{Infotainment} topics is removed from the graph, while \textit{Performance} is split into two separate topics. Furthermore, the \textit{Handling} topic was removed. As the coherence score increases with $k$, topics remaining in \Cref{tab:topics-muse} (\graph, $k$ = 2) have a higher topic coherence score than the ones removed.

\begin{table}[t]
\centering
\resizebox{.48\textwidth}{!}{%
\begin{tabular}{ccc}
\hline
\hline
\textbf{Inferred Topic}   & \textbf{Topic Representatives}   & \textbf{Gold Topic} \\ \hline
\hline
 & \textbf{K-Means}    & \\
\hline
\textit{Handling} & suspension, handling, dampers, corners, chassis   & Handling \\
\textit{Infotainment} & menus, satnav, swivel, commands, entertainment & User Experience  \\
\textit{Interior Features} & dash, design, events, wood, plastic & Interior Features \\
\textit{Performance} & engine, turbo, litre, cylinder, engines & Performance  \\
\textit{Safety}  & detection, assist, safety, collision, airbags   & Safety  \\
\textit{Storage} & storage, items, space, boot, hooks & General Information \\
\textit{YouTube} & please, enjoy, click, share, wow & General Information \\
\textit{Miscellaneous} & cars, guys, opportunity, brand, tomorrow       & General Information \\
\hline

   & \textbf{GraphTMT} (K= 1)   &  \\ \hline
\textit{Infotainment} & navigation, controls, touch, apple, buttons & User Experience  \\
\textit{Infotainment} & hand, pop, screen, entertainment, information & User Experience  \\
\textit{Passenger Space}  & area, head, roof, room, headroom   & Interior Features  \\
\textit{Handling} &  suspension, corners, steering, gear, response & Handling \\
\textit{Performance} & seconds, turbo, twin, acceleration, cylinder       & Performance \\
\textit{YouTube} & channel, dot, please, thanks, share & General Information \\
\hline

   & \textbf{GraphTMT} ($k$ = 2)&  \\ \hline
\textit{Infotainment}  & hand, pop, screen, entertainment, information   & Infotainment  \\
\textit{Passenger Space} &  seat, back, headroom, room, head & Handling \\
\textit{Performance} & seconds petrol miles diesel economy gallon fuel & Performance  \\
\textit{Performance} & seconds, turbo, acceleration, twin, cylinder & Performance  \\
\textit{YouTube} & dot, channel, please, wow, share & General Information \\        
\end{tabular}%
}
\caption{List of topics extracted on \musec{} where K-Means uses TF-weighted; \graph{} uses \pthresh{} = 0.8. \label{tab:topics-muse}}
\end{table}

\section{\city{} Evaluation}

In the second part of our evaluation, we compare the performance of all four models on the \city{} to show \graph{}'s applicability outside of YouTube transcripts. The models are compared on their coherence score, topic coverage, and topic overlap. %Note that the Citysearch vocabulary includes all part-of-speech words.

\subsection{Coherence Score Comparison}

\Cref{tab:cityPerformance} lists the results of the best performing models based on their coherence scores. K-Means and \graph{} ($k$=3) result in the highest coherence score, and \lda{} has the lowest. Similar to \musec{}, K-Means gets better scores using TF instead of TF-IDF and increasing $k$ has a positive effect on the coherence score of \graph{} but reduces the number of topics.
\city{} has six gold topics, but K-Means infers eight and \graph{} ($k$=3) results in five topics. At $k$ = 1 our approach infers nine topics but has a lower score than \lda{}. \hdbscan{} performed similar to K-Means but infers only three topics. 

These scores show that our approach is applicable outside of YouTube transcripts, achieving the highest $c_v$ score. Furthermore, they confirm a previous finding, increasing $k$ results in a better score but fewer topics.

%\begin{tabular}[c]{@{}c@{}}$\alpha$ = 1/$T_n$,\\ $\beta$ = 0.4, \\ $T_n$ = 8\end{tabular} &
%\begin{tabular}[c]{@{}c@{}}$k$ = 8,\\ TF-weighted\end{tabular}  & 
%\begin{tabular}[c]{@{}c@{}}$k$ = 1,\\ \pthresh = 0.8\end{tabular}  &

\begin{table}[t!]
\centering
\resizebox{.8\linewidth}{!}{% .5\textwidth
\begin{tabular}{l|c|c|cc}
\hline
\textbf{Topic Models}  & $\bm{T_n}$   & $\bm{c_v}$   &  $\bm{T_{c}}$   &   $\bm{T_{o}}$  \\ \hline

\lda($\alpha$ = 1/$T_n$, $\beta$ = 0.40)   & 8 & .48   & .67 & .50           \\

K-Means(TF-weighted)  & 8 & \textbf{.64}  & \textbf{.83} & .38 \\ 

\hdbscan($C_{min}$=5) & 3 & .61 & .33 & .33 \\ 

\graph{} ($k$ = 1, \pthresh = 0.80) &  9  & .40  & .67 & .56 \\ 

\graph{} ($k$ = 2, \pthresh = 0.80) &  6  & .60  & .67 & .33  \\ 

\graph{} ($k$ = 3, \pthresh = 0.80) &  5  & \textbf{.64}  & .67 & \textbf{.20}

\end{tabular}%
}
\caption{Results on the \city{} for four different topic models  (\lda, K-Means, \hdbscan, \graph) and three metrics: coherence score ($c_v$), topic coverage ($T_c$), and topic overlap ($T_o$). \label{tab:cityPerformance}}
%\vspace{-2em}
\end{table}

\subsection{Intra-Topic Assessment}

The previous scores show that K-Means and \graph{} ($k$=3) have the best overall topic coherence. In the following, we look at their topic coverage and overlap (cf. \Cref{tab:musecPerformance}). \Cref{tab:topics-citysearch} lists all K-Means topics, their inferred labels, and the model's gold topic coverage. The table shows that K-Means covers five of the six gold topics ($T_c$ = .83): \textit{Ambience}, \textit{Anecdotes}, \textit{Food}, \textit{Miscellaneous},\textit{Price}, but \textit{Anecdotes}, and \textit{Food} are captured twice ($T_o$ = .375).

All \graph{} models cover four of the six gold topics but as $k$ increases, the topic overlap decreases. \Cref{tab:topics-citysearch} lists the nine \graph{} ($k$=1) topics, their inferred labels, and the topic coverage. Comparing these topics with the topics at $k$=3 shows the effect of $k$ on \graph. Increasing the edge connectivity parameter lowers the number of topics but can also let new topics turn up (\ie \textit{Ambient} is in \graph{} ($k$=3) but not in \graph{} ($k$=1)). This shows that topics can hold more semantics than indicated by their representatives, and increasing $k$ can split an existing topic into semantically different topics, showing the hierarchical structure of our graph-based approach. 
%These results indicate that the approach can be applied to other datasets.

%%%%%%%%%%%%%%%%%%%%%%
%%%%%%%%%%%%%%%%%%%%%%
%
%
%
% Discussion
%
%
%%%%%%%%%%%%%%%%%%%%%%
%%%%%%%%%%%%%%%%%%%%%%
\section{Discussion \label{sec:discuss}}

\begin{table}[t]
\centering
\resizebox{.48\textwidth}{!}{%
\begin{tabular}{ccc}
\hline
\hline
\textbf{Inferred Topic}   & \textbf{Topic Representatives}   & \textbf{Gold Topic} \\ 
\hline
\hline

   & \textbf{K-Means}   &  \\ \hline
\textit{Ambience} & comfy, spacious, calm, sleek, couch &  Ambience \\
\textit{Miscellanous} & appear, control, clue, sight, fooled & Miscellanous  \\
\textit{Anecdotes} & yesterday, today, tonight, march, celebrate & Anecdotes \\
\textit{Anecdotes} & refused, proceeded, busboy, ignored, annoyed & Anecdotes  \\
\textit{Price}  & normal, pay, normally, expensive, afford  & Price  \\
\textit{Location} & south, astoria, williamsburg, ues, houston & Miscellaneous \\
\textit{Food} & yogurt, pear, pate, walnut, cinnamon & Food \\
\textit{Food} & sliced, char, pate, prawn, chorizo & Food \\
\hline

 & \textbf{GraphTMT}  ($k$ = 1) &  \\
\hline
\textit{Food}         & pickled, seed, puree, fennel, curried           & Food       \\
\textit{Food}         & poivre, hanger, hangar, flank, frites           & Food       \\
\textit{Service (neg.)}           & unhelpful, unattentive, unapologetic, arrogant, unfriendly             & Staff             \\
\textit{Service (pos.)}          & responsive, cordial, polite, gracious, professional        & Staff            \\
\textit{Location}               & washington, seaport, murray, madison, greene              & Miscellaneous \\
\textit{Location}           & chelsea, downtown, soho, meatpacking, tribeca                 & Miscellaneous       \\
\textit{Location}           & brand, england, yorker, orleans, yorkers                      & Miscellanous       \\
\textit{Anecdotes}               & incredible, outstanding, terrific, excellent, fantastic           & Anedcote                 \\
\textit{Time}               & tuesday, wednesday, monday, friday, thursday              & Anedcote \\
\hline

     & \textbf{GraphTMT}  ($k$ = 3)                                  & \textbf{Gold Topic} \\ \hline
\textit{Food}         & pickled, seed, puree, fennel, curried           & Food       \\
\textit{Service (neg.)}           & unhelpful, unattentive, unapologetic, arrogant, unfriendly            & Staff             \\
\textit{Service (pos.)}          & responsive, engaging, sincere, caring, hospitable      & Staff            \\
\textit{Anecdotes}               & flavorless, tasteless, overcooked, undercooked, inedible           & Anecdotes               \\
\textit{Ambient}               & painted, tile, lantern, banquette, chandelier              & Ambient              
\end{tabular}%
}
\caption{List of topics extracted on \city{} where K-Means uses TF-weighted; \graph{} uses \pthresh{} = 0.8. \label{tab:topics-citysearch}}
\end{table}
We evaluated the competitiveness of our novel graph-based topic modeling approach to common alternatives (\lda, K-Means, \hdbscan) on two different datasets (\musec, \city). 
% coherence score
Our experiments have shown that \graph{} achieves the highest coherence scores on \musec{} and \city{}. Furthermore, the model's edge-connectivity parameter ($k$) positivly affects the coherence score but decreases the number of topics. These findings suggest that by varying $k$ we can remove incoherent topics and words that do not semantically align with a topic. We should note that K-Means had the same coherence score on \city{} but with more topics. All other models (\lda, \hdbscan) scored less on both datasets. Although K-Means achieved a comparable score on \city{} with more topics, the model requires one to predefine the number of topics. 
% As  \graph{} does not require the true number of topics, it is a good alternative for when this information is not available.
Since \graph{} does not require a specification of the (true) number of topics, it is a good alternative if this information is not available, should not be predetermined, or a search for a suitable parameter $k$ can not be performed. Moreover, the automatic retrieval of $k$ by techniques such as the elbow method is controversial and rarely optimal~\cite{ketchen1996application}.

% study
In addition to comparing the semantic coherence of topics, we conducted a user study to assess the human interpretability of the \musec{} topics. The study included the models with the highest coherence scores (\lda, K-Means, \graph). As in previous studies, the resulting coherence scores align with the coherence scores~\cite{roderCoherence15}, \graph{} topics were more interpretable than topics from K-Means and \lda. 

% intra-topic assessment
The intra-topic assessment allowed us to compare topics from K-Means and \graph, the two highest scoring models on both datasets. K-Means covered more gold topics, but \graph{} resulted in topics with less overlap. Note that varying $k$ revealed the hierarchical structure of \graph, increasing the parameter can split a topic into two semantically different topics.

These findings suggest that  \graph{} provides a valid alternative to common topic model techniques as users can interpret the topics better, more unique topics are extracted, and the approach  does not require the true number of topics. Overall, this study has shown the relevance of graph connectivity in topic modeling on two different datasets (YouTube transcripts and online restaurant reviews). %In the future, we want to compare different graph connectivity algorithms (\eg{} clique percolation method) to find and develop even more effective approaches for topic extraction.

% In this study, we preprocessed automatic transcripts using a POS tagger to extract all nouns. This approach removes any unknown words appearing in the transcripts due to the word error rate. 
In our experiments, \graph{} has proven to be very robust on a spoken dataset with a high word error rate. We want to validate these findings on other datasets in future work. Furthermore, we want to evaluate different preprocessing approaches for transcript. Another future aim is to compare different graph connectivity algorithms (\eg{} clique percolation method) to find and develop even more effective approaches for topic extraction.

\section{Conclusion}
In this paper, we demonstrated the capability of graph-based topic modeling on real-world YouTube transcribed data (\musec) and textual reviews (\city). On the \musec{} dataset, our proposed novel \graph{} outperforms all three baseline models in terms of cluster coherence, uniqueness, and interpretability. An accompanying user study assessed the last one. On the \city{} dataset, our method achieves competitive results to K-Means. However, the clusters produced by \graph{} have less semantic overlap. We conclude that graph-based clustering is a valid alternative for topic modeling on transcripts and provides meaningful results on real-world text datasets. For the future, we will focus on an integrated approach of several modalities, such as, vision, audio and metadata as any attempt at drawing meaning from YouTube must consider all aspects.

\bibliographystyle{unsrt}
\bibliography{refs}

\end{document}